\pdfoutput=1

\documentclass[11pt]{article}

\usepackage[final]{acl}

\usepackage{times}
\usepackage{latexsym}

\usepackage[T1]{fontenc}

\usepackage[utf8]{inputenc}
\usepackage{todonotes}

\usepackage{microtype}

\usepackage{inconsolata}

\usepackage{wrapfig}
\usepackage{array}

\usepackage{amsmath}
\usepackage{comment}
\usepackage{dialogue}
\usepackage{mdframed}
\usepackage{environ}
\usepackage{algorithm}
\usepackage{algpseudocode}
\usepackage{amssymb} 
\usepackage{xcolor}
\usepackage[italian,english]{babel}

\newcommand{\hdialo}{Dial$_\text{H}$}
\newcommand{\hllmdialo}{Dial$_\text{H+LLM}$}
\newcommand{\llmdialo}{Dial$_\text{LLM}$}

\newcommand{\editPedix}{$_{\textit{p-e}}$}
\newcommand{\origPedix}{$_{\textit{orig}}$}

\newmdenv[linecolor=black, linewidth=1.5pt,innerlinewidth=0.5pt, leftmargin=30pt, rightmargin=10pt]{dialoguebox}

\newcommand{\prompttable}[4]{%
    \begin{table}[h!]
        \small
        \centering
        \begin{tikzpicture}
        \node (table) [inner sep=0pt] {
        \begin{tabular}{p{0.82\linewidth}}\
        \\
        \textbf{\textsc{Prompt #1}} \\
        #2 \\
        \\
        \end{tabular}
        };
        \draw [rounded corners=.5em, very thick] (table.north west) rectangle (table.south east);
        \end{tikzpicture}
        \caption{#3}
        \label{tab:#4}
    \end{table}
}

\NewEnviron{dialoguetable}[4]{%
  \begin{table}[h!]
    \begin{tabular}{p{#1\linewidth} p{#2\linewidth}}
      Speaker & Turn \\ \hline
      \BODY
    \end{tabular}
    \caption{#3}
    \label{tab:#4}
  \end{table}
}

\usepackage{multirow}
\usepackage{booktabs} 
\usepackage{rotating}
\usepackage{enumitem}
\usepackage{tabularray}
\usepackage{tikz}

\definecolor{correctionGreen}{RGB}{34,139,34}

\title{Fine-tuning with \texttt{HED-IT}: \\ The impact of human post-editing for dialogical language models}

\author{
 Daniela Occhipinti$^{1,2}$,
 Michele Marchi$^{1,2}$,
 Irene Mondella$^{3,4}$,
 Huiyuan Lai$^3$, \\
\textbf{ Felice Dell'Orletta$^4$,
 Malvina Nissim$^3$,
 Marco Guerini$^1$}
 \\
 $^1$Fondazione Bruno Kessler, Italy, 
$^2$University of Trento, Italy\\
 $^3$University of Groningen, Netherlands, 
 $^4$ItaliaNLP Lab @ CNR-ILC, Italy\\
\texttt{\{docchipinti, mmarchi, guerini\}@fbk.eu, i.mondella@studenti.unipi.it.} \\
\texttt{felice.dellorletta@ilc.cnr.it, \{h.lai, m.nissim\}@rug.nl}
}

\begin{document}
\maketitle
\begin{abstract}

Automatic methods for generating and gathering linguistic data have proven effective for fine-tuning Language Models (LMs) in languages less resourced than English. 
Still, while there has been emphasis on data quantity, less attention has been given to its quality. In this work, we investigate the impact of human intervention on machine-generated data when fine-tuning dialogical models. In particular, we study (1) whether post-edited dialogues exhibit higher perceived quality compared to the originals that were automatically generated; (2) whether fine-tuning with post-edited dialogues results in noticeable differences in the generated outputs; and (3) whether post-edited dialogues influence the outcomes when considering the parameter size of the LMs.
To this end, we created \texttt{HED-IT}, a large-scale dataset where machine-generated dialogues are paired with the version post-edited by humans. Using both the edited and unedited portions of \texttt{HED-IT}, we fine-tuned three different sizes of an LM. Results from both human and automatic evaluation show that the different quality of training data is clearly perceived and it has an impact also on the models trained on such data. Additionally, our findings indicate that larger models are less sensitive to data quality, whereas this has a crucial impact on smaller models. These results enhance our comprehension of %
the impact of human intervention on training data in the development of high-quality LMs.
\end{abstract}

\begin{figure}[!t]
\centering
\includegraphics[width=1\columnwidth]{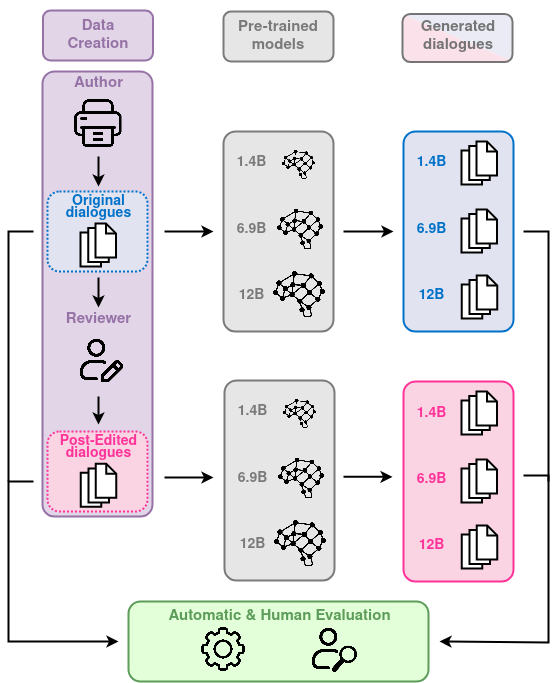}
 \caption{Data creation pipeline. The \textit{author} (machine) provides initial dialogues, the \textit{reviewer} (human) post-edits them. These two datasets are then used to train different sizes of an LM. Automatic and human evaluation is performed both on training data and corresponding trained models to assess the impact of curation.}
 \label{fig:pipeline_summary}
 \end{figure}

\section{Introduction}
The landscape of language technology is constantly being populated by new %
instruction-based and conversational models, %
thanks to the increased availability of foundational Large Language Models (LLMs) and easy access to fine-tuning support. %
Especially for lesser-resourced languages, the standard procedure is to automatically translate existing English datasets %
and use those to fine-tune one of the available LMs on the target language (see \citet{naveed2023comprehensive} for a recent overview of models and procedures).

Still, this proliferation of models is not matched by progress in evaluation efforts aimed at understanding exact differences, especially in terms of data and model quality as perceived by humans (rather than just overall model performance).
Gaining such insights would help in optimising the balance between cheaper and more costly choices, such as pure automatic translation vs manual intervention, in the creation of datasets for developing high-quality models. This is the aim of the present contribution, focusing on Italian dialogic models, but applicable to other under-resourced languages. By building on previous research revolving around the concept of data quality, %
we adopted an \textit{author-reviewer} architecture \cite{tekiroglu-etal-2020-generating} to create the \texttt{HED-IT} (Human Edited Dialogues for ITalian) dataset\footnote{The dataset is available for research purposes at \url{https://github.com/LanD-FBK/hed-it}.} where automatically generated dialogues are post-edited by humans. We then use such edited dialogues, as well as the unedited, to fine-tune an LLM in three different sizes. This setup enables us to study the impact of human intervention on training data %
(vs its absence) on the final resulting models. We do so in terms of (i) quality assessment by humans of LLM output; and (ii) behaviour assessment across model sizes. %
Figure~\ref{fig:pipeline_summary} provides an overview of our approach. The research questions posed and addressed in this paper can be summarised as follows:

\begin{description}[leftmargin=*]
\itemsep0em

\item[Q1] We investigate whether post-edited dialogues to be used for fine-tuning are (perceived as) better in quality than the original (automatically generated) ones. Automatic metrics and dedicated manual evaluation suggest that this is the case.  

\item[Q2] We investigate if fine-tuning with post-edited data yield detectable differences in the fine-tuned models. %
Model evaluation via automatic metrics and output evaluation via human judgement indicate an overall quality improvement for the models fine-tuned with post-edited data.

\item[Q3] We study the impact of the LLM's size and observe that the impact of post-edited data is stronger for smaller model sizes.

\end{description}

\noindent In addition to providing a dataset of more than 16,000 original and post-edited Italian dialogues, and a battery of fine-tuned models of different sizes, we contribute insights and a methodology that can be applied to other languages and tasks.

\section{Related Work}

\paragraph{LLMs for Conversational Models}
LLMs with billions of parameters, pre-trained on a huge amount of language data in a self-supervised manner,  achieved great success 
~\citep{brown2020language, naveed2023comprehensive},
especially as the backbone of conversational agents 
often achieving human-like interaction~\citep{thoppilan2022lamda, openai-chatgpt, wei2023leveraging}. However, no open-source model, especially in languages other than English, comes close in capabilities to closed-source models, and creating instruction-following data for fine-tuning can be highly time-consuming and costly. This has prompted research into finding ways to reduce such efforts. For instance, \citet{wang-etal-2023-self-instruct} proposed a framework to fine-tune LLMs using instructional signals from the model itself, reducing the need of human-written instructions. Furthermore, knowledge distillation~\citep{hinton2015distilling}, which involves transferring knowledge from larger to smaller models, has been used to train open-source conversational agents~\citep{xu-etal-2023-baize, vicuna2023}. Still, instruction-based or conversational models in languages other than English are typically trained using automatically translated data~\citep{chen2023monolingual, larcher2023cabrita, ranaldi2023empowering}. This data, mostly in the form of instructions, originates in English. For instance, a suite of LLaMA-based models in various languages has been trained using the same dataset that was initially developed and utilised for English Alpaca~\citep{alpaca}. %
In this work, we specifically focus on the Italian landscape.

\paragraph{Italian Conversational Models}
The Italian model corresponding to English Alpaca~\citep{alpaca} is Camoscio \citep{santilli2023camoscio}, which is trained with low-rank adaptation (LoRA, \citealt{hu2022lora}) on the Alpaca dataset automatically translated to Italian. Stambecco\footnote{\url{https://github.com/mchllabs/stambecco}; no other documentation is available for this model.} is an Italian instruction-following model, based on LLaMA~\citep{touvron2023llama1} and fine-tuned on the Italian translation of GPT-4-LLM dataset~\citep{peng2023instruction}.
Fauno~\citep{bacciu2023fauno} is a more conversation-like Italian model, trained on %
English data translated to Italian, originally generated by ChatGPT chatting with itself. Most recently, \citet{basile2023llamantino} released LLaMAntino-Chat, a family of models also based on the LLaMA~2 Chat \cite{touvron2023llama2}, adapted on the Italian language by fine-tuning on the Italian translation of the UltraChat dataset \cite{ding2023enhancing}.%

 \paragraph{Low-Resource Languages Dialogue Datasets}
 Several approaches have been adopted in literature to collect multilingual or low-resource language datasets. Various methodologies automatically translate English data to the language of interest \cite{bacciu2023fauno, santilli2023camoscio, chen2023monolingual, larcher2023cabrita, ranaldi2023empowering}. In their work, \citet{10.1162/tacl_a_00609} propose the Multi$^3$WOz dataset, a comprehensive multilingual, multi-domain, task-oriented dataset with dialogues in four languages: English, Arabic, French, and Turkish. These dialogues were obtained using a Wizard of Oz (WoZ) methodology, where bilingual speakers, native in the target language and fluent in English, crafted natural and culturally relevant conversations. However, this method was resource-intensive, consuming significant human and financial resources and taking over a year to compile the dialogues. Another dialogic dataset, both in English and Chinese, is BiToD \cite{lin2021bitod}. BiToD is a task-oriented dataset in which dialogues are collected through a dialogue simulator interacting with a knowledge base to generate dialogue outlines that are then revised by crowdsourcers to convert them into natural conversations. While this method resembles our proposed approach, the goals differ significantly. \citet{lin2021bitod} are interested in the bilingual and cross-lingual evaluation of the task. In contrast, we focus on in-depth analyses of a low-resource language and open-domain dataset with original and revised versions, assessing data quality, collection efficiency, and their effects on model fine-tuning across different sizes.

\paragraph{Data Quality}
The quality of fine-tuning plays a crucial role in improving performance with less data ~\citep{zhou2023lima}. While the benefits of post-editing in machine translation are long-known 
\citep{castilho-etal-2014-post, castilho-obrien-2016-evaluating}, in developing LLMs the main focus has been on data curation (document filtering) to increase quality ~\citep{zhou2023oasis, li2023quantity, wang2024survey, furman-etal-2023-high}, while less attention has been given to document correction via post-editing \cite{tekiroglu-etal-2022-using}.

\section{Dataset Creation}

To create the \texttt{HED-IT} dataset, we used the \textit{author-reviewer} pipeline proposed by \citet{tekiroglu-etal-2020-generating}, in which an LLM (the \textit{author} component) produces novel data, while humans (the \textit{reviewer} component) filter and eventually post-edit them. We tested several author configurations and introduced two adaptations to the original proposal: (i) following \citet{russo-etal-2023-countering}, rather than fine-tuning a GPT model, %
we opted for an instruction-based LLM (i.e., ChatGPT \cite{openai-chatgpt}) that does not require fine-tuning; (ii) we extend the concept of a machine-based \textit{author} in this setting to include also the automatic extraction of excerpts from existing human-produced text.  %
To create the dataset, we used non-real data sourced from movie scripts and synthetic data to address potential privacy and ephemerality concerns associated with real data %
\cite{klubicka2018examining}.%

The following sections provide a detailed overview of the sources and methodologies used to construct the \texttt{HED-IT} dataset.

\subsection{Author Module}
For the author module, we tested three strategies, whose common feature is that they do not rely on the manual creation of new dialogues from scratch, which is particularly costly: (i) automatically extracting excerpts from human-written material, %
(ii) using an LLM to rewrite human-written dialogues, (iii) automatically generating dialogues prompting an LLM with human-written material.
Our aim in using these configurations was to create a dataset with a wide range of linguistic characteristics and topics. We constrained the number of speakers to two and the number of turns to a minimum of three to maintain clarity and consistency, avoiding complex multi-character conversations. 

\paragraph{Human Dialogues (\hdialo)}%

To obtain human-written dialogues without having to produce them anew, we exploited movie scripts scraped from various sources, with no genre-specific filters. In this case, the author module is an algorithm that extracts script portions meeting some given characteristics (e.g. excerpts between two speakers, minimum three turns, further details are in Appendix~\ref{sec:appendix-human-dialogues}).

\paragraph{Human-LLM Dialogues (\hllmdialo)}%
One limitation of extracting portions of movie dialogues is that, by nature, they rely heavily on visual support, which can make them less understandable when taken out of context. %
To reintroduce some contextual depth to the extracted dialogues, and to make them self-contained, we instructed ChatGPT to rewrite them in a more comprehensible way %
(details and prompt in Appendix~ \ref{sec:review_human_dialogues}).%

\paragraph{LLM Dialogues (\llmdialo)} 

Gathering new Italian dialogues can be costly, and utilizing real data poses potential privacy issues. Thus, following the approach suggested by \citet{chen-etal-2023-places}, we generated synthetic dialogues by instructing %
ChatGPT with diverse prompts using various contexts (further details in~Appendix~\ref{sec:context_dialogue_prompt}). %
The contexts provided in the prompt were obtained from four distinct sources:
(i) a set of background information about two speakers and a topic sourced from PLACES \cite{chen-etal-2023-places}, %
automatically translated from English to Italian using DeepL; %
(ii) a set of conversation-starting questions obtained from a website which supports social interaction; %
(iii) a set of tweets from the X/Twitter account of ANSA\footnote{\url{https://twitter.com/Agenzia_Ansa}}, one of the main news agencies in Italy; %
(iv) a subset of the Twitter Italian Dialect Data dataset\footnote{\url{https://www.kaggle.com/datasets/alonyoeli/twitter-italian-dialect-data}}.%

\subsection{Reviewer Module}

Two Italian native speakers (MSc students in Computational Linguistics) were hired as annotators for the post-editing process. Following the methodology proposed by \citet{fanton-etal-2021-human}, both annotators underwent extensive training on the data and task, including initial sessions to establish guidelines. In these sessions, they were given a selection of author-generated dialogues to identify common issues %
and to propose the most effective strategies for %
enhancing their quality. Subsequently, these strategies were integrated into the guidelines. Once the annotation campaign started, weekly meetings were held to address any possible issues or feedback %
from annotators. 

Post-editing guidelines aimed to reduce annotators' effort %
while maintaining the quality of the output. %
Below, we discuss examples of common phenomena and post-editing actions, while in Table \ref{tab:postediting-example}, we present an example of dialogue post-editing.

\begin{table}[h!]
    \small
    \centering
    \begin{tikzpicture}
        \node (table) [inner sep=0pt] {
            \begin{tabular}{p{0.01\linewidth}p{0.40\linewidth}|p{0.40\linewidth}}
                \\
                 & \textbf{\textsc{Original}} & \textbf{EN}\\
                A: & \foreignlanguage{italian}{\textcolor{red}{Mettetevi in circolo.} Lei chi è?} & \textcolor{red}{Make a circle.} Who are you?\\
                B: & \foreignlanguage{italian}{Sono il marito.} & I am the husband.\\
                A: & \foreignlanguage{italian}{Spiacente, soltanto le persone \textcolor{red}{interessate} sono ammesse qui.} & I am sorry, only \textcolor{red}{interested} people are allowed here.\\
                &&\\
                 & \textbf{\textsc{Post-Edited}} & \textbf{EN} \\
                A: & Lei chi è? &  Who are you?\\
                B: & Sono il marito. & I am the husband.\\
                A: & Spiacente, soltanto le persone \textcolor{correctionGreen}{coinvolte nel caso} sono ammesse qui. & I am sorry, only people \textcolor{correctionGreen}{involved in the case} are allowed here.\\
                \textcolor{correctionGreen}{B:} & \textcolor{correctionGreen}{Ma io sono suo marito! Non mi potete negare l'accesso.} & \textcolor{correctionGreen}{But I am her husband! You cannot negate me the access.} \\
                \\
            \end{tabular}
        };
        \draw [rounded corners=.5em, very thick] (table.north west) rectangle (table.south east);
    \end{tikzpicture}
    \caption{Example of dialogue post-editing. In \textcolor{red}{red} the deletions and in \textcolor{correctionGreen}{green} the additions.}
    \label{tab:postediting-example}
\end{table}

\paragraph{Post-Editing of \hdialo}

Excerpts extracted from movie scripts present unique phenomena due to their specific linguistic characteristics, including dialect usage, visual references, and fictional situations common in genres like horror, fantasy, and sci-fi.
Therefore, tailored guidelines are necessary: %
\begin{itemize} [leftmargin=*]
\itemsep0em
    \item Overly generic or difficult-to-understand dialogues, such as \foreignlanguage{italian}{\textit{Beh, e... / Se ne hai voglia, ne posso trovare quanta ne vuoi. / Come?}} (EN: \textit{Well, and... / If you feel like it, I can find as much as you want. / How?}), requiring excessive effort for post-editing, should be removed.
    \item Sometimes, dialogues may contain dialectal terms. If the terms are comprehensible to those unfamiliar with the dialect used, they can be kept, such as the Tuscan \foreignlanguage{italian}{\textit{O questo chi l'è?}} (EN: \textit{Who is this?}). Otherwise, they should be removed.
    \item Interjections typical of the spoken language (e.g. \textit{Ah!}, \textit{Mmm...}, \textit{Eh!}) can be kept.
    \item To maintain coherence, overly long dialogues or including abrupt topic changes should be split.
    \item Dialogues may have unclear or unhelpful turns, which should be removed for clarity. If a turn is deleted, either the preceding or following turn must be removed or merged to maintain the alternating speakers.
    \item Sometimes, dialogues may lack initial context or a proper ending. In such instances, annotators can create and insert a beginning and/or concluding turn to ensure coherence in the final dialogue.
\end{itemize}

\paragraph{Post-Editing of \hllmdialo{} and \llmdialo{}}

Specific guidelines were necessary for post-editing dialogues either rewritten or fully generated by ChatGPT. %
Key issues include repetitive and redundant content, as well as excessive use of politeness characterized by frequent agreement and continuous expressions of gratitude. This tendency towards politeness might be attributed to the use of %
Reinforcement Learning With Human Feedback \cite{NEURIPS2020_1f89885d}, which encourages extremely cautious and gentle responses, often resulting in unrealistic dialogues. Even when instructed to express strong disagreement, the model tends to produce excessives level of politeness.
\begin{itemize} [leftmargin=*]
\itemsep0em
    \item Repetitive phrases such as \foreignlanguage{italian}{\textit{"Capisco il tuo punto di vista, ma..."}} (EN: \textit{"I understand your point of view, but..."}) and \foreignlanguage{italian}{\textit{"Non capisco come tu possa..."}} (EN: \textit{I don't understand how you can...}) should be either rephrased or deleted.
    \item Generated dialogues can include redundant turns or information, e.g., 
    \foreignlanguage{italian}{\textit{Su questo siamo d'accordo. Grazie per la tua opinione, Marco. / Di nulla, Alice. È sempre interessante discutere di questi temi con te.}} (EN: \textit{We agree on this. Thank you for your opinion, Marco. / You're welcome, Alice. It is always interesting to discuss these topics with you.}).  
    This content should be deleted.
    \item Generated dialogues sometimes appear excessively polite, with frequent agreement and continuous expressions of gratitude, %
    such as \foreignlanguage{italian}{\textit{Sì, hai ragione. Ma.. / Sì, forse hai ragione. Ma... / Assolutamente d'accordo. Ma...}} (EN: \textit{Yes, you are right. But... / Yes, you may be right. But... / Absolutely agree. But...})
    Such instances should be curbed as much as possible.
    \item The endings of the generated dialogues often appear repetitive and standardised. These closures should be reworded as much as possible.
\end{itemize}

\section{Dataset Analysis}
Following the author-reviewer process, two versions of our dataset were collected: one with the original dialogues (\texttt{HED-IT}\origPedix{}) and another with the post-edited versions (\texttt{HED-IT}\editPedix{}). \texttt{HED-IT}\origPedix{} comprises over 9000 dialogues and more than 66000 turns, whereas \texttt{HED-IT}\editPedix{}  comprises over 7000 dialogues and almost 48000 turns. On average, dialogues contain 8 turns. More details are reported in Table~\ref{tab:dataset_metrics_1}.
\begin{table}[ht!]
\small
\centering
\begin{tabular}{p{1.52cm}|r|r|r|r}

\toprule

                                     & \textbf{\hdialo} & \textbf{\hllmdialo} & \textbf{\llmdialo}    & \textbf{Total} \\ \midrule
         Dial\origPedix{}     & 2504           & 3458               & 3339            & 9301           \\
                                     Dial\editPedix{} & 1107           & 2824               & 3266            & 7197           \\ \midrule
              Turns\origPedix{}    & 14042          & 22214              & 30034           & 66290          \\
                                     Turns\editPedix{}   & 7817           & 16422              & 23704           & 47943          \\ \midrule \midrule
Turns/Dial\origPedix{}     & 6              & 7                 & 12               & 8            \\
                                     Turns/Dial\editPedix{}    & 8              & 7                 & 10               & 8           \\ \midrule 
   Tok/Dial\origPedix{} & 138            & 164               & 424              & 251            \\
                                 Tok/Dial\editPedix{}   & 173            & 140               & 332              & 232  \\ \midrule
 Tok/Turn\origPedix{}     & 24              & 24               & 37               & 31     \\
                                 Tok/Turn\editPedix{}    & 23              & 22               & 35               & 29     \\ \bottomrule
\end{tabular}
\caption{Dimensions of the \texttt{HED-IT}\origPedix{} and \texttt{HED-IT}\editPedix{} datasets in terms of dialogues and turns. The upper part reports counts, while the lower reports averages.}
\label{tab:dataset_metrics_1}
\end{table}

To assess the validity of the methodology employed to build \texttt{HED-IT}, we also conducted an additional analysis where a human annotator created a set of dialogues from scratch. We then compared the time needed for post-editing a set of dialogues versus composing them from the beginning. The results are summarised in Table \ref{tab:scratch_vs_pe}.

\begin{table}[ht!]
    \centering
    \small
    \begin{tabular}{l|r|r}
    \toprule
         & \textbf{From scratch} & \textbf{Post-editing} \\
    \midrule
         Dial. per hour & 17  & \textbf{39} \\
         Turns per hour & 158  & \textbf{381} \\
         Tok. per hour 	& 1695 & \textbf{6627}\\
    \bottomrule
    \end{tabular}
    \caption{Time needed (in seconds) for dialogues written from scratch by a single human annotator vs. machine-generated dialogues post-edited by the same annotator.}
    \label{tab:scratch_vs_pe}
\end{table}

The table demonstrates a marked productivity gain between dialogues written by a human annotator and those that are machine-generated and then post-edited by the same annotator. The latter results in more than double the dialogues and turns processed, and nearly four times the tokens, highlighting a significant efficiency improvement, crucial for dataset scalability\footnote{A similar improvement using an \textit{author-reviewer} pipeline is reported in \cite{russo-etal-2023-countering}.}. This is particularly substantial if we compare our proposed methodology to the WoZ scenario as in Multi$^3$WOz, where not only dialogues are written from scratch, but also each simulated turn causes one participant to wait in an "idle state" for the other to answer, theoretically doubling the annotator's required time.

To assess the impact of post-editing both quantitatively and qualitatively, we use automatic metrics and a human evaluation.

\subsection{Automatic Metrics}

Inspired by \citet{russo-etal-2023-countering}, we evaluated the quality of the post-edited data compared to the original data employing the \textit{Human-targeted Translation Edit Rate} (HTER; \citet{snover-etal-2006-study}) and the \textit{Repetition Rate} (RR; \citet{bertoldi-etal-2013-cache}).

\paragraph{HTER}
measures the number of editing required to transform a machine-generated text into its post-edited version. The score is computed considering word substitution, insertions and deletions.

\paragraph{RR} is used to assess how repetitive a text is. It calculates the geometric mean of the frequency of repeated sequences of words (\textit{n-grams}) within the text. To ensure consistency across texts of varying lengths, a fixed-size sliding window is used during processing. In our analysis, we focused on word \textit{n-grams}, where \textit{n} varies from 1 to 4, and employed a sliding window of 1000 words. \\ %

Table \ref{tab:dataset_metrics_2} summarises the impact of human intervention across the different portions of the dataset. Several phenomena can be seen.   %
(i) %
\hdialo{} have a higher deletion rate than other authors' output, both in terms of whole dialogues (0.559 vs 0.191 and 0.022) and turns  (0.436 vs 0.249 and 0.186). A post-hoc interview with the annotator showed that this is mainly due to their being ``decontextualized excerpts" and over-reliant on visual context, making them unsuitable for simple post-editing. %
(ii) In contrast, %
\hllmdialo{} and \llmdialo{} tend to have fewer deletions. This implies that more dialogues and turns are post-edited (0.373 vs 0.761 and 0.954). Similarly, the percentage of post-edited turns is higher. %
Interestingly, the required post-editing is typically less intensive (HTER of 0.616 vs HTER of 0.358). (iii) However, we also observed that \llmdialo{} %
have a more ``artificial" nature: there is an increase in RR as compared to other author configurations (4.771 vs 2.603 and 2.467). (iv) Finally, there is a clear beneficial effect of the post-editing, that consistently enhances the quality of dialogues by reducing the RR in all configurations.

\begin{table}[ht!]
\small
\centering
\begin{tabular}{l|r|r|r|r}
\toprule
& \multicolumn{1}{l|}{\textbf{\hdialo{}}} & \multicolumn{1}{l|}{\textbf{\hllmdialo{}}} & \multicolumn{1}{l|}{\textbf{\llmdialo{}}} & \multicolumn{1}{l}{\textbf{Total}} \\ \midrule
Dial$_{\textit{unch}}$ & \textbf{0.068} & 0.048 & 0.024 & 0.045 \\
Dial$_{\textit{del}}$ & 0.559 & 0.191 & \textbf{0.022} & 0.229 \\ 
Dial$_{\textit{edit}}$ & 0.373 & 0.761 & \textbf{0.954} & 0.726 \\ \midrule
Turns$_{\textit{unch}}$ & 0.475 & 0.480 & \textbf{0.524} & 0.501 \\
Turns$_{\textit{del}}$ & 0.436 & 0.249 & \textbf{0.186} & 0.253 \\ 
Turns$_{\textit{edit}}$ & 0.089 & 0.271 & \textbf{0.290} & 0.253 \\ \midrule
HTER$_{\textit{edit}}$ & 0.616 & 0.611 & \textbf{0.358} & 0.459 \\ \midrule
RR\origPedix{}  & 2.603 & \textbf{2.467} & 4.771 & 3.790 \\
RR\editPedix{}  & 2.428 & \textbf{1.992} & 3.590 & 2.965 \\ 
\bottomrule
\end{tabular}
\caption{Analysis of Post-Editing interventions. \textit{unch}: turn/dialogue was not changed;  \textit{del} turn/dialogues was removed during post-editing; \textit{edit}: intervention that was not deletion. HTER is computed only on edited turns.  %
}
\label{tab:dataset_metrics_2}
\end{table}

\subsection{Human Evaluation}\label{sec: human evaluation}
Besides the automatic metrics, we used human evaluation to assess the impact of post-editing.
We selected %
a total of 416 dialogues from the %
\llmdialo{} subset: 208 post-edited dialogues (half by one annotator  %
and half by the other) and their corresponding original versions. %
Dialogues from movie scripts and the version rewritten by the author model were excluded  %
as they may be easily recognisable, leading to possible  %
bias into the responses.

The sample was evaluated by native Italian speakers recruited via Prolific\footnote{\url{https://www.prolific.com/}}, a platform specifically designed for research purposes. %
Each evaluator received a survey consisting of 8 dialogues stratified by length: \textit{short dialogues} with 10 or fewer turns, \textit{medium} with between 10 and 15 turns and \textit{long} with more than 15 turns.  
To cover the 416 dialogues, we created 52 surveys, each assigned to 3 different evaluators. In total, we had almost 2500 evaluations from 129 different participants\footnote{Some participants completed more than just one survey.}.%

The surveys %
aimed to assess the dialogues along two dimensions: understandability and naturalness (defined as the "likelihood that it was written by a machine"), both using a 5-point scale.
The responses were categorised into evaluations regarding the comprehensibility and naturalness of both the Dial\origPedix{} %
and Dial\editPedix{} %
dialogues. The mean response was then calculated for each subgroup (micro-average). Results %
are reported in \autoref{tab:dataset_human_evaluation}: the mean Understandability of original dialogues is lower than for post-edited dialogues (4.045 and 4.175 respectively), while Machine Probability is higher (3.296 vs 2.875). This shows that post-editing leads to more understandable and less artificial dialogues. %
Furthermore, comparing each original dialogue directly with its post-edited version and calculating the average responses of every dialogue, shows  %
 0.130 increase in understandability and a 0.421 decrease in artificiality, on average.

\begin{table}[ht!]
\centering
\resizebox{\columnwidth}{!}{
\begin{tabular}{ll|r|r}
\toprule & & \textbf{Understand}~$\uparrow$ & \textbf{Mach. Prob.}~$\downarrow$ \\ \midrule
\multirow{3}{*}{\textbf{orig}} & Low edit & 4.122 & 3.160 \\ 
& High edit & 3.968 & 3.433 \\ \cmidrule{2-4}
& Overall & 4.045 & 3.296 \\\midrule
\multirow{3}{*}{\textbf{p-e}} & Low edit & 4.266 & 2.913 \\ 
& High edit & 4.083 & 2.837\\ \cmidrule{2-4}
& Overall & \textbf{4.175} & \textbf{2.875} \\\bottomrule
\end{tabular}
}
\caption{Human evaluation answers' micro-averages.}
\label{tab:dataset_human_evaluation}
\end{table}

Dialogues were also sorted by post-editing rate, dividing them into \textit{highly post-edited} and \textit{lowly post-edited}. To do so, we took into account the HTER and the number of eliminated turns for each dialogue. \autoref{tab:dataset_human_evaluation} shows also that, for the second question (Machine Probability), the difference between original and post-edited is greater for highly post-edited dialogues than for lowly post-edited ones (a reduction of 0.596 vs a reduction of 0.247). This is caused by the fact that the highly post-edited dialogues were perceived as very unnatural in their original version, and therefore benefited greatly from post-editing (the more post-editing, the greater the improvement in naturalness).

\section{Experimental Design}

Using the \texttt{HED-IT} dataset, our objective is to understand the impact of post-editing on the performances of an LLM. We evaluated the models automatically using the test set gold standard, and ran human evaluations of models' outputs.

\subsection{Dataset}

To obtain two datasets of equal sizes, we started from the
\texttt{HED-IT}\editPedix{}, which has fewer dialogues compared to \texttt{HED-IT}\origPedix{}, and split it into train, validation and test subsets (80:10:10), stratified according to author-strategy 
(i.e., \hdialo{}, \hllmdialo{}, and \llmdialo). We then subsampled the same number of dialogues from the \texttt{HED-IT}\origPedix{} partition including original dialogues that had been deleted during post-editing. In this way, we both controlled for the training data size and granted that the distribution of subsampled data represents the distribution of the real \textit{Original} data, i.e. including low-quality dialogues as well. After the subsampling, 83\% of the original dialogues were matched with their post-edited counterpart, while the remaining 17\% came from the ``deleted" set.

\subsection{Models}

As base models, we use Pythia \cite{biderman2023pythia}, a suite of decoder-only autoregressive language models. Pythia's LLMs are trained on publicly available data, maintaining the same order of training data and varying only in size from 70M to 12B. This setup allows us to isolate the impact of post-editing on the LLMs' generations, considering the LLMs dimensions as the only variable\footnote{While other LMs are available for Italian that could be used for fine-tuning, Pythia is the only one that allows us to run controlled experiments including size, and rule out the effect of other variables.}.

For this study, we focused on the Pythia sizes of 1.4B, 6.9B and 12B. We conducted a preliminary experiment to assess the proficiency of these LLMs in Italian in a zero-shot setting, which showed that these models are suitable for our experiments. Subsequently, we fine-tuned these models with QLoRA \cite{dettmers2023qlora} in six different configurations. These configurations involved each of the three Pythia model sizes, fine-tuned on either the \textit{Original} and the \textit{Post-Edited} dialogues (see Figure~\ref{fig:pipeline_summary}). Furthermore, we ensured that the models were fine-tuned with the same data order, consistently with Pythia's methodology. %
For details on fine-tuning configurations, see Appendix \ref{fine-tuning-config}.

\subsection{Model Evaluation}

To assess the quality of the models we used the following automatic metrics:
\begin{itemize} [leftmargin=*]
    \item \textbf{Conditional turn Perplexity (CPPL)} \cite{su-etal-2021-put, occhipinti2023prodigy} 
    measures the perplexity of a gold turn given the previous dialogue history. %
    We computed the CPPL for each turn iteratively and considered the average.
    \item \textbf{Average Accuracy at N (Acc@N)} \citet{welch-etal-2022-leveraging} 
    defines 
    the prediction of a word from a gold turn being correct if it appears within the top \textit{N} most probable words provided by the model. We computed the Acc@N at each turn incrementally, considering the average across all turns.

\end{itemize}

\noindent Models fine-tuned on original data were tested on original data, while those fine-tuned on post-edited data were tested on post-edited data.

\subsection{Output Generation}

To let the models generate dialogues we prompt them with the first two turns of a subset of  100 dialogues which have been used for the human evaluation of the dataset (Section~\ref{sec: human evaluation}). For generation, we used the \textit{top-p} decoding mechanism with a value of 0.9, a temperature of 1, and a repetition penalty of 2. As we use six models, we obtain a total of 600 generated dialogues.

We noticed a consistent pattern in all model-generated dialogues: since we forced the model to generate content extensively, eventually, it reached a limit where it began repeating turns or using synonyms excessively. This resulted in unnatural and inconsistent dialogue.
We call this phenomenon ``derailment". To automatically detect the start of derailment in each model's output, we used BLEU \cite{papineni2002bleu} with a threshold of 0.9 to identify when the turns become overly similar. Table~\ref{tab:generations-hum-eval-stats} shows %
the average length of dialogues and turns when the derailment is cut off.
Models trained on post-edited dialogues appear to derail sooner compared to models trained on original dialogues. This could be attributed to the length discrepancy between post-edited and original dialogues: during fine-tuning, models trained on post-edited data are exposed to shorter dialogues (251 vs 232 token on average per dialogue, see Table~\ref{tab:dataset_metrics_1}). Consequently, at inference time, they might find it challenging %
to generate dialogues as lengthy as those produced by models trained on original dialogues. 

\begin{table}[ht!]
\small
\centering
\begin{tabular}{l|r|r|r}

\toprule
\textbf{Model}                & \multicolumn{1}{r|}{\textbf{Turns}} & \textbf{Tok/Dial} & \textbf{Tok/Turn} \\ \toprule
\textbf{1.4B\origPedix{}} & 14 & 267 & 19 \\
\textbf{1.4B\editPedix{}} & 13 & 251 & 19 \\ \midrule
\textbf{6.9B\origPedix{}} & 13 & 394 & 31 \\
\textbf{6.9B\editPedix{}} & 12 & 314 & 27 \\ \midrule
\textbf{12B\origPedix{}}  & 21 & 420 & 20 \\
\textbf{12B\editPedix{}}  & 14 & 402 & 29 \\ \bottomrule
\end{tabular}
\caption{Average lengths of generated dialogues considering BLEU score 0.9 derailment cut. }
\label{tab:generations-hum-eval-stats}
\end{table}

\subsection{Output Evaluation}

\noindent Human evaluation was carried out in the same manner as for the dataset analysis, namely with crowdsourced assessments over understandability and probability of automatic generation (Section \ref{sec: human evaluation}). Derailment turns were excluded, and dialogues were cut at 20 turns to prevent negative impacts from derailment and excessive length.

\section{Results}

In this section, we will outline the results obtained on the six configurations we tested.

\subsection{Automatic Evaluation}

The CPPL scores in Table~\ref{tab:cppl-values}  show that models trained on post-edited dialogues consistently outperform those trained on original dialogues across all Pythia dimensions. Notably, the CPPL difference between models trained on original and post-edited dialogues is lower for the biggest model (a difference of 2.462 for 1.4B vs a difference of 0.748 for 12B), 
indicating a more significant impact of post-editing on smaller models. %

\begin{table}[ht!]
\small
\centering
\begin{tabular}{l|rr}
\toprule
\textbf{Model}                & \textbf{CCPL} & \textbf{$\Delta$}  \\ \midrule
\textbf{1.4B\origPedix{}} & 14.016    & \multirow{2}{*}{2.462} \\
\textbf{1.4B\editPedix{}} & 11.554    &                        \\ \midrule
\textbf{6.9B\origPedix{}} & 13.098    & \multirow{2}{*}{2.663} \\
\textbf{6.9B\editPedix{}} & \textbf{10.465}    &                        \\ \midrule
\textbf{12B\origPedix{}}  & 11.733    & \multirow{2}{*}{0.748} \\
\textbf{12B\editPedix{}}  & 10.985    &                        \\ \bottomrule
\end{tabular}
\caption{Models' CPPL values over gold turns.}
\label{tab:cppl-values}
\end{table}

\begin{table*}[ht!]
    \small
    \centering
    \begin{tabular}{l|rrr|rrr|rrr}
        \toprule
        \textbf{} & \multicolumn{3}{c|}{\textbf{All turns}} & \multicolumn{3}{c|}{\textbf{-20\%}} & \multicolumn{3}{c}{\textbf{-30\%}} \\
        \textbf{Model}& \textbf{Acc@10} & \textbf{Acc@5} & \textbf{Acc@1} & \textbf{Acc@10} & \textbf{Acc@5} & \textbf{Acc@1} & \textbf{Acc@10} & \textbf{Acc@5} & \textbf{Acc@1} \\
        \midrule
        \textbf{1.4B\origPedix{}} & 0.823 & 0.769 & 0.583 & 0.823 & 0.767 & 0.582 & 0.818 & 0.763 & 0.578 \\
        \textbf{1.4B\editPedix{}} & 0.817 & 0.761 & 0.571 & 0.821 & 0.765 & 0.578 & 0.818 & 0.763 & 0.577 \\
        \midrule
        \textbf{6.9B\origPedix{}} & 0.855 & 0.805 & 0.618 & 0.854 & 0.804 & 0.616 & 0.850 & 0.799 & 0.612 \\
        \textbf{6.9B\editPedix{}} & 0.850 & 0.798 & 0.608 & 0.855 & 0.803 & 0.616 & 0.852 & 0.800 & 0.615 \\
        \midrule
        \textbf{12B\origPedix{}} & \textbf{0.863} & \textbf{0.815} & \textbf{0.627} & \textbf{0.862} & \textbf{0.811} & \textbf{0.625} & \textbf{0.858} & \textbf{0.807} & \textbf{0.621} \\
        \textbf{12B\editPedix{}} & 0.852 & 0.802 & 0.612 & 0.857 & 0.806 & 0.620 & 0.855 & 0.804 & 0.619 \\
        \bottomrule
    \end{tabular}
    \caption{Models' Acc@N over gold turns,  with 20\% and 30\% of dialogues' last turns progressively removed.}
    \label{tab:accn}
\end{table*}

Table~\ref{tab:accn} presents Acc@N scores for various configurations. Acc@N values are higher for models trained on original dialogues compared to post-edited dialogues across all turns (e.g., 0.627 of Acc@1 for 12B\origPedix{} vs 0.612 of Acc@1 for 12B\editPedix{}). We hypothesise that the presence of excessively polite responses generated by the Author model (ChatGPT) in original dialogues, often edited out or rephrased in post-edited versions, may contribute to this difference, especially considering the prevalence of \hllmdialo{} and \llmdialo{} dialogues. To test this hypothesis, we recomputed the metrics systematically excluding the last 20\% and 30\% of each test set dialogue. We indeed observed a decrease in Acc@N for original dialogues, leading to a convergence in performance between original and post-edited dialogues (e.g. 0.621 of Acc@1 for 12B\origPedix{} vs 0.619 of Acc@1 for 12B\editPedix{}).

\begin{table}[ht!]
\centering
\small
\begin{tabular}{l|rr|rr}
\toprule
     & \multicolumn{2}{c|}{\textbf{Original}} & \multicolumn{2}{c}{\textbf{Post-Edited}} \\ \midrule
     & pre-cut & post-cut & pre-cut & post-cut \\ \midrule
\textbf{1.4B} & 1.888 & 1.534            & 1.651 & \textbf{1.434}             \\
\textbf{6.9B} & 2.637 & 2.022            & 2.149 & \textbf{1.705}             \\
\textbf{12B}  & 1.928 & \textbf{1.490}            & 2.202 & 1.781              \\ \bottomrule
\end{tabular}
\caption{Repetition Rate (RR) over generated dialogues.}
\label{tab:RR-generations}
\end{table}

The output of the models has been automatically assessed using the RR metric. The results are presented in Table \ref{tab:RR-generations}. The scores demonstrate that removing the derailment consistently improves RR scores in all configurations. Moreover, results show that the impact of the post-editing is more visible in smaller models. In fact, smaller models trained on post-edited dialogues have lower RR rates compared to those trained on original dialogues (e.g., 1.434 for the 1.4B model trained on post-edited dialogues vs 1.534 for the same model trained on original dialogues). Conversely, for the largest models, the RR is higher for the model trained on original dialogues compared to its counterpart trained on post-edited dialogues (i.e., 1.490 vs 1.781).

\subsection{Human Evaluation}

For the human evaluation of the generated dialogues, we selected 600 dialogues (100 for each model), resulting in 75 different surveys submitted for crowdsourced ratings. Each dialogue is assessed by three participants. Participants from the first human evaluation (Section~\ref{sec: human evaluation}) were excluded from this round.
Examples of well and poorly rated dialogues are given in \autoref{appendix:human_evaluation_generations}. In total, we had  3600 evaluations from 169 different participants.

Results are in \autoref{tab:human_evaluation}. Models trained on post-edited dialogues generate text perceived as more natural, as the Machine Probability score for these models is always lower than for models trained on the original dialogues (4.137 of 1.4b\editPedix{} vs 4.237 of 1.4b\origPedix{}, 3.887 vs 4.037 for the 6.9b model and 3.927 vs 4.077 for the 12b one). Understandability improves with post-editing only for the largest model among those considered (3.493 of 12b\editPedix{} vs 3.447 of 12b\origPedix{}), while the other two models do not seem to improve when trained on post-edited dialogues. In summary, humans perceive the impact of post-editing training data on model generation. %

\begin{table}[ht!]
    \small
    \centering
    \begin{tabular}{l|c|c}
        \toprule
          &  \textbf{Understand} $\uparrow$ &  \textbf{Machine Prob.} $\downarrow$\\\midrule
         \textbf{1.4b\origPedix{}} &  2.987 &  4.237\\
         \textbf{1.4b\editPedix{}} & 2.977 & 4.137\\\midrule
         \textbf{6.9b\origPedix{}} &  3.430 &  4.037 \\
         \textbf{6.9b\editPedix{}} & 3.377 & \textbf{3.887}\\\midrule
         \textbf{12b\origPedix{}} &  3.447&   4.077 \\
         \textbf{12b\editPedix{}} & \textbf{3.493} & 3.927\\\bottomrule
    \end{tabular}
    \caption{Human evaluation of generated dialogues.} %
    \label{tab:human_evaluation}
\end{table}

\section{Conclusions}

Machine-generated data is commonly used to fine-tune LLMs in low-resourced languages, while improving data quality has received little attention. To this end, we extensively investigated the impact of human intervention on data used for fine-tuning dialogical models. We provided a corpus of machine-generated Italian dialogues paired with post-edited counterparts, and trained an LLM in three different sizes using this corpus. Results clearly showed that post-edited dialogues had a higher perceived quality compared to their original counterparts. Model evaluation also indicates an overall quality improvement for the models fine-tuned with post-edited data. Still, results suggest that the effect of post-editing is stronger for smaller models while slowly fading for bigger ones, and this can be a critical aspect when running a data collection campaign. Also the intervention of machine-rewriting before human post-editing can be beneficial when available training data, even if human-written, is silver with respect to the target domain.

\section*{Limitations}

This work has a few limitations worth noting. Firstly, the resource presented is available only in Italian. %
However, it is important to highlight that the author-reviewer approach used for data collection is language-independent and can be replicated assuming that an LLM for the target language is available. 

A significant portion of the data was generated by prompting an LLM (ChatGPT): it is important to recognise that prompting autoregressive models can bring to uncontrolled forms of generation, such as hallucinations that for some scenarios might be critical. 

Additionally, it is important to acknowledge that our testing was limited to only a subset of possible models (i.e. Pythia), although we would not expect great differences in results with other models. 

Finally, while human post-editing can greatly improve data quality, it might require substantial resources in terms of human effort, time, and financial investment if author module's outputs are not good enough.

\section*{Ethics Statement}

In this study, two of the authors acted as annotators in the post-editing process. For human evaluation, we used crowdsourcing via the Prolific platform. All crowdsourced workers were compensated following the platform's ethical payment principles\footnote{\url{https://researcher-help.prolific.com/hc/en-gb/articles/4407695146002-Prolific-s-payment-principles}}. Finally, it is essential to highlight that the study did not assess the safety of the models used, but our fine-tuning data were all validated by the annotators, so any safety concern is mainly due to pre-training data.

\bibliography{custom}

\clearpage

\appendix
\section{Author Component}
\subsection{Human Excerpts Extraction Algorithm}\label{sec:appendix-human-dialogues}
Algorithm \ref{alg:sliding_window} outlines the process for extracting human excerpts from movie scripts. The method involves maintaining a minimum window size, checking how many authors are within it, and determining if the dynamic window could be expanded further. Initially, if the starting window contains more than two actors, it is shifted by one unit; otherwise, it is enlarged until it meets a third actor, at which point the excerpt without the third actor is saved. After saving, the window returns to its minimum size and put in such a way that there cannot be overlapping turns with already saved dialogues. A final check (lines 19-20) is performed to save the last portion of an excerpt if it meets the specified criteria. The implementation of \texttt{SAVE\_SUB\_DIALOGUE} can be customized as required. For our use case we utilized a minimum window size of 3.
\begin{algorithm}
\small
\begin{algorithmic}[1]
    \Function{dynamicSlidingWindow}{$scene, window\_size$}
        \State $begin \gets 1$
        \State $end \gets window\_size$
        
        \While{$end  \leq (\text{length of } scene)$}
            \If{\Call{unique}{$scene.author[begin:end]$})$ > 2$}
                \State $end \gets end - 1$
                \If{$end - begin < window\_size$}
                    \State $begin \gets begin + 1$
                    \State $end \gets begin + window\_size$
                \Else
                    \State \Call{save\_sub\_dialogue}{$scene[begin:end]$}
                    \State $begin \gets end + 1$
                    \State $end \gets begin + window\_size$
                \EndIf
            \EndIf
            \State $end \gets end + 1$
        \EndWhile

        \State $end \gets end - 1$
        \If{\Call{unique}{$scene.author[begin:end]$})$ = 2$ \textbf{and} $end - begin \geq window\_size$}
            \State \Call{save\_sub\_dialogue}{$scene[begin:end]$}
        \EndIf
        
    \EndFunction
\end{algorithmic}
\caption{Dynamic Sliding Window}\label{alg:sliding_window}
\end{algorithm}

\subsection{Prompt for \hllmdialo{} generation}\label{sec:review_human_dialogues}

\hllmdialo{} has been generated by prompting ChatGPT with the input specified in Table \ref{tab:PromptMovieChatGPT}. While we present the prompt in English for ease of understanding, we utilised its Italian version in the actual interactions. The dialogues were generated using top-p decoding with a value of 0.9 and a temperature of 0.8 via OpenAI APIs.

\prompttable{ }{%
Rewrite the following dialogue in its entirety in a way that is meaningful, natural, realistic, coherent, comprehensible and self-conclusive.\\
\{\{DIALOGUE\}\}}{Prompt to rewrite excerpts extracted from movies with ChatGPT.}{PromptMovieChatGPT}

\subsection{Prompt for \llmdialo{} generation}\label{sec:context_dialogue_prompt}

To identify the most effective instruction, we conduct preliminary experiments involving testing various prompts. Annotators evaluated the quality of the generated data using a sample, based on factors like variability, originality, and coherence of the generations. In one experiment, we instructed the model to generate dialogues solely using contextual information. In another, we instructed it to first establish the personalities of two speakers based on the context, then generate the entire dialogue considering both the context and the speakers' personalities. We observed that explicitly requesting distinct personalities resulted in more varied dialogues, whereas without such instruction, the generated dialogues were similar for style and structure. Therefore, we provided the model with a specific context and instructed it to develop the personalities of two speakers engaging in a dialogue about that context. An example of prompt is provided in Table \ref{tab:contextPrompt}. Although we display the prompts in English for clarity, the actual interactions took place in Italian. Also in this case, the dialogues were generated using OpenAI APIs, employing \textit{top-p} decoding with parameters set to 0.9 and a temperature of 0.8.

\prompttable{1}{Given the text that follows {>}{>}{>}, come up with two characters connected to it and describe their personality. Make then a dialogue between the two that is natural, realistic, coherent, comprehensible and self-contained. In the dialogue, there can be only the two actors and their turns. The two actors do not necessarily have to agree with each other. The dialogue must not be artificial and excessively friendly.\\
\\
The output structure is:\\
\\
Character description: \\
\\
Speaker1: description\\
Speaker2: description\\
\\
Dialogue:\\
\\
Speaker1: turn\\
Speaker2: turn \\
\\
{>}{>}{>} [[CONTEXT]]}{Example of prompt used to generate dialogues with ChatGPT, translated from Italian.}{contextPrompt}

\section{Reviewer Component: Detailed Guidelines}

In this section, we outline the methodology guidelines employed to review our dataset's dialogues. The main focus is to post-edit the Italian dialogue so to ensure that they appear natural, self-contained, and clear without being vague while minimising the manual intervention. Each dialogue must consist of at least three turns, with exactly two speakers taking alternating turns.

\subsection*{Optimal Dialogues and Turns}

\begin{itemize}
    \item Preferred dialogues and turns should be natural, realistic, contextually appropriate, and easily understandable.
    \item \textbf{Italian dialects}: Italian dialects can be included briefly, but the main message should still be clear to readers unfamiliar with those dialects. Since Italian is mostly based on Tuscanian, this dialect is generally more tolerated.

\item \textbf{Non-fluent expressions}: expressions such as \textit{Ha}, \textit{Ah} and \textit{Oh} can be kept, although commonly associated with spoken language. 

\item \textbf{Expression related to communication mediums:} this kind of expressions, such as speaking into a radio, can be kept.

\item \textbf{Common typos}: typos that are commonly associated with keyboard limitations are acceptable. However, if the same typos consistently appear in all turns, it is recommended to correct them occasionally (e.g. using \textit{E’} instead of \textit{È}). 

\end{itemize}

\subsection*{Sub-optimal Dialogues and Turns}

\begin{itemize}
    \item \textbf{Dialogues with less than three turns}: if a dialogue consists of fewer than three turns,  either the dialogue can be deleted or new turns can be added to ensure it has at least three turns;
    \item  \textbf{Excluding non-Italian languages}: if a movie script excerpt contains passages in languages other than Italian, or if ChatGPT occasionally generates non-Italian dialogues, those non-Italian dialogues or turns should be removed. This rule also applies in-universe (e.g. fantasy) languages.

    \item \textbf{Extensive use of Italian dialects}: if a dialogue uses Italian dialects extensively, it becomes harder to read and understand. Therefore, such dialogue should be removed.

\item \textbf{One-way communication}: This refers to a situation where one party in a conversation is merely providing verbal or non-verbal cues of attention or approval, without contributing any meaningful context or information. These dialogues should be post-edited or removed.

\item\textbf{Distorted Italian}: Some characters might speak Italian in a distorted manner, influenced by their native language or origins. This distortion could involve using incorrect grammar, speaking only in infinitives, or mixing in excessive jargon from their original language (e.g``ogre speaking''). Dialogues or turns with this characteristic should be post-edited for clarity or, if necessary, removed.

\item \textbf{Non-Italian onomatopoeic expressions}:  onomatopoeic expressions not commonly found in Italian, such as \textit{Euh}, can be removed. 
\item \textbf{Excessively vague dialogues}: sometimes, dialogues can be overly vague, lacking clear meaning or purpose. This typically occurs in short exchanges. Such dialogues should be removed.

\item\textbf{Sung Parts:} Some dialogues contain sung sections, usually in English and recognizable by the use of the “/” character. These sections should be removed.

\item \textbf{Individual speech}: sometimes, even if in a dialogue, characters might speak without replying to what the other is saying. This occurs, for instance, when characters are in conflict or when there is a speaker device like a radio or television involved. These instances should either be removed or adjusted during post-editing.

\item \textbf{Non-verbal interruptions}: these interruptions occur in spoken dialogues when one speaker uses non-verbal cues to signal agreement or encourage the other speaker to continue, resulting in the interruption of the current speaker's turn. These interruptions can be eliminated and the turns of the interrupted speaker can be merged.

\item \textbf{Grammatical errors}: when a dialogue or a turn contains grammatical errors (e.g. {dì} instead of \textit{di'} for meaning \textit{dire}, \textit{qual'è} instead of \textit{qual è}, \textit{qualcun'altro} instead of \textit{qualcun altro}), they must be fixed.
\item\textbf{References to external context}:  If external content is frequently mentioned, especially through expressions like \textit{qui} or \textit{là}, consider removing such references.
\item \textbf{Excessive politeness in dialogues}: Sometimes, ChatGPT generates dialogues that are excessively polite, with frequent agreement and continuous expressions of gratitude, resulting in unnatural conversations. Unnecessarily polite phrases can be removed. If an entire dialogue is excessively polite, it may be deleted.

\end{itemize}

\subsection*{How to perform turn deletion}
Dialogues must consist of two alternating speakers. Therefore, any deletions in the middle of the conversation must be done in pairs. This means either removing one speaker's turn and merging it with the previous or next turn of that same speaker, or removing two turns together. The only exception to the pair deletion rule is when the turn slated for deletion is at the beginning or end of the dialogue. In such cases, a deletion may involve only one turn.

When a speaker in a dialogue delivers two consecutive turns, those turns should be merged into a single turn.

\subsection*{How to perform turn addition}

Adding turns to a dialogue can significantly improve its clarity and coherence, particularly when there is a lack of context. We recommend only to add turns at the beginning or at end of a dialogues, as adding them midway requires a comprehensive understanding of the context prior and after the section in which is planned an addition, beyond the need to design an even number of turns so to keep the alternation of speaker, requirement which are considered excessively demanding effort-wise.

\subsection*{How to modify dialogues by merging or splitting}

If there are two consecutive dialogues that appear as a continuation of each other, these can be merged.

In case there is a dialogue with multiple occurring contexts (e.g. when characters are discussing one topic and then something significant happens that shifts the focus). If both parts still meet the basic criteria of a valid dialogue (at least 3 exchanges that are clear and self-conclusive), they can be separated.

\section{Experimental Details}
\subsection{Fine-Tuning Configuration}\label{fine-tuning-config}

Each Pythia model dimension was fine-tuned employing QLoRA \cite{dettmers2023qlora}, with low-rank approximation set to 64, low-rank adaptation set to 16, and dropout rate set to 0.1 Evaluation steps were set at 720, batch size at 4 and gradient accumulation step at 2. We employed 50 epochs with early stopping after 5 epochs. The original learning rates from \citet{biderman2023pythia} were utilised. The final epochs and learning rates are detailed in Table \ref{tab:hyperparameters}. For the fine-tuning process, an Nvidia Ampere A40 GPU with 48GB memory was utilised.

\begin{table}[h]
\centering
\small
\begin{tabular}{l|r|r}
\toprule
\textbf{Model} & \textbf{Epochs} & \textbf{Learning Rate} \\
\midrule
\textbf{1.4B\origPedix{}} & 11  & $2.0 \times 10^{-4}$ \\
\textbf{1.4B\editPedix{}} & 10  & $2.0 \times 10^{-4}$ \\
\textbf{6.9B\origPedix{}} & 10  & $1.2 \times 10^{-4}$ \\
\textbf{6.9B\editPedix{}} & 9  & $1.2 \times 10^{-4}$ \\
\textbf{12B\origPedix{}} & 19 & $1.2 \times 10^{-4}$ \\
\textbf{12B\editPedix{}} & 13 & $1.2 \times 10^{-4}$ \\
\bottomrule
\end{tabular}
\caption{Models' fine-tuning epochs and learning rates}
\label{tab:hyperparameters}
\end{table}

\subsection{Automatic Results Details}

\subsubsection{CPPL Details}
Table \ref{tab:cppl-values-no-last-turns} displays CPPL values with 20\% and 30\% of the last turns in dialogues progressively excluded. It demonstrates that models trained on post-edited data consistently outperform those trained on original data, even when the last turns of the dialogues are removed. Generally, removing the last turns improves model performance in terms of CPPL, indicating a negative impact of these turns on model performance. %
\begin{table}[ht!]
    \small
    \centering
    \begin{tabular}{l|rrr}
        \toprule
        \textbf{Model} & \textbf{All turns} & \textbf{-20\%} & \textbf{-30\%} \\
        \midrule
        \textbf{1.4B\origPedix{}} & 14.016 & 13.543 & 14.188 \\
        \textbf{1.4B\editPedix{}} & 11.554 & 11.369 & 11.638 \\
        \midrule
        \textbf{6.9B\origPedix{}} & 13.098 & 10.511 & 11.011 \\
        \textbf{6.9B\editPedix{}} & \textbf{10.465} & 8.301 & 8.474 \\
        \midrule
        \textbf{12B\origPedix{}} & 11.733 & 10.024 & 10.422 \\
        \textbf{12B\editPedix{}} & 10.985 & \textbf{8.064} & \textbf{8.215} \\
        \bottomrule
    \end{tabular}
    \caption{Models’ CPPL values over gold turns, with 20\% and 30\% of dialogues' last turns progressively removed.}
    \label{tab:cppl-values-no-last-turns}
\end{table}

\section{Human Evaluation}\label{appendix:human_evaluation_generations}
Example of dialogue negatively evaluated (Understandability of 1 and Machine Prob. of 4.667):
\begin{table}[H]
    \small
    \centering
    \begin{tikzpicture}
        \node (table) [inner sep=0pt] {
            \begin{tabular}{p{0.01\linewidth}p{0.40\linewidth}|p{0.40\linewidth}}
                \\
                 & \textbf{\textsc{IT}} & \textbf{EN}\\
                A: & Ciao Giulia, come stai? Spero che tu possa aiutarmi in questa situazione difficile. & Hi Giulia, how are you? I hope you can help me in this difficult situation.\\
                B: & Ciao Luca, certo, dimmi pure qual è la tua situazione attuale. & Hi Luca, sure, please tell me what is your current situation.\\
                A: & Grazie per essere qui con me, mi hai raccontato tutto quello che sto cercando di fare. & Thank you for being here with me, you have told me everything I am trying to do.\\
                B: & Non c'è bisogno di spiegarmi, sei un grande amico. &  No need to explain, you are a great friend. \\
                A: & Lo so, grazie ancora Giulia. & I know, thank you again Giulia. \\
                B: &  È stato un piacere parlare con te. Buona giornata! & It's been a pleasure talking with you. Have a great day! \\
                A: &  Grazie, a presto! & Thank you, see you soon! \\
                B: &  Ciao! & Bye! \\
                \\

            \end{tabular}
        };
        \draw [rounded corners=.5em, very thick] (table.north west) rectangle (table.south east);
    \end{tikzpicture}
    \caption{Example of a negatively evaluated dialogue.}
    \label{tab:neg_dialogue}
\end{table}

Example of dialogue positively evaluated (Understandability of 4.667 and Machine Prob. of 1.333):
\newpage

\begin{table}[H]
    \small
    \centering
    \begin{tikzpicture}
        \node (table) [inner sep=0pt] {
            \begin{tabular}{p{0.02\linewidth}p{0.40\linewidth}|p{0.40\linewidth}}
                \\
                 & \textbf{\textsc{IT}} & \textbf{EN}\\
                A: & Ciao Carla, come va? & Hi Carla, how are you doing? \\
                B: & Bene grazie, e tu? & Fine thank you, how about you? \\
                A: & Tutto bene, grazie. Volevo chiederti una cosa. & I'm all right, thank you. I wanted to ask you something. \\
                B: & Certo, dimmi pure. &  Sure, tell me. \\
                A: & Vorrei sapere se ti piacerebbe venire a cena con me domenica prossima. &  I would like to know if you would like to have dinner with me next Sunday. \\
                B: &   Sì, sarebbe fantastico! &  Yes, that would be great! \\
                A: &   Perfetto, allora ci vediamo domenica prossima. & Perfect, then I'll see you next Sunday \\
                B: &  Fantastico! A domenica! &  Great! See you on Sunday! \\
                A: &  A domenica! & See you on Sunday! \\
                \\

            \end{tabular}
        };
        \draw [rounded corners=.5em, very thick] (table.north west) rectangle (table.south east);
    \end{tikzpicture}
    \caption{Example of a positively evaluated dialogue.}
    \label{tab:pos_dialogue}
\end{table}

\end{document}